\pdfoutput=1

\documentclass[11pt]{article}

\usepackage[]{acl}

\usepackage{times}
\usepackage{latexsym}

\usepackage{amsmath}
\usepackage{amssymb}
\usepackage{mathtools}
\usepackage{amsthm}

\usepackage{times}
\usepackage{latexsym}

\usepackage[utf8]{inputenc} 
\usepackage[T1]{fontenc}    
\usepackage{hyperref}       
\usepackage{url}            
\usepackage{booktabs}       
\usepackage{amsfonts}       
\usepackage{nicefrac}       
\usepackage{microtype}      
\usepackage{epsfig,graphicx,subfigure}
\usepackage{multirow,booktabs,hhline}
\usepackage[algo2e]{algorithm2e}
\usepackage{bm}
\usepackage{wrapfig}
\usepackage{lipsum}
\usepackage{footnote}
\usepackage{bbding}
\usepackage[bottom]{footmisc}
\usepackage[toc,page]{appendix}
\usepackage{color}         

\usepackage[capitalize,noabbrev]{cleveref}

\usepackage[T1]{fontenc}

\usepackage[utf8]{inputenc}

\usepackage{microtype}

\title{Toward Building General Foundation Models for Language, Vision, and Vision-Language Understanding Tasks}

\author{Xinsong Zhang\thanks{\ \ Correspondence to: <xszhang0320@gmail.com>.}  \\
  ByteDance Research \\   
  \And
  Yan Zeng \\
  ByteDance Research \\ \And
  Jipeng Zhang \\
  HKUST \\ \And
  Hang Li \\
  ByteDance Research \\
  }

\newcommand{\ModelName}{X-FM\xspace}
\newcommand{\ModelNameB}{X-FM$_\mathrm{base}$\xspace}

\newcommand{\baby}{X$^2$-VLM\xspace}
\newcommand{\babyx}{X$^2$-VLM}

\begin{document}
\maketitle

\begin{abstract}
Foundation models or pre-trained models have substantially improved the performance of various language, vision, and vision-language understanding tasks. However, existing foundation models can only perform the best in one type of tasks, namely language, vision, or vision-language. 
It is still an open question whether it is possible to construct a general foundation model performing the best for all the understanding tasks. In this paper, we propose a new method for training the general foundation model, {\ModelName} (the X-Foundation Model).  {\ModelName} has one language encoder, one vision encoder, and one fusion encoder, as well as a new training method. The training method includes two new techniques for learning {\ModelName} from text, image, and image-text pair data. One is to stop gradients from the vision-language training when learning the language encoder. The other is to leverage the vision-language training to guide the learning of the vision encoder. Extensive experiments on benchmark datasets show that {\ModelName} can significantly outperform existing general foundation models and perform better than or comparable to existing foundation models specifically for language, vision, or vision-language understanding. 
Code and pre-trained models are released at \url{https://github.com/zhangxinsong-nlp/XFM}.

\end{abstract}

\setlength{\parskip}{0.55ex}
\vspace{-0.3em}
\section{Introduction}
\label{sec:intro}
With the enormous power of foundation models, also known as pre-trained models, remarkable performance gains have recently been achieved in a variety of understanding tasks in natural language processing (NLP), computer vision (CV), and other fields~\citep{devlin2018bert,liu2019roberta,lewis2019bart,raffel2020exploring,brown2020language,dosovitskiy2020image,he2022masked,bao2021beit,lu2019vilbert,tan-bansal-2019-lxmert,chen2020uniter,li2020oscar,li2021align,zeng2021multi,zeng2022x}
. Foundation models are usually equipped with Transformer~\citep{vaswani2017attention} as the backbone, pre-trained with a tremendous amount of unlabeled data, and then fine-tuned with small amounts of labeled data in downstream tasks. The strong representation ability of the model, the massive amount of data, and the effective means of training make the foundation models powerful for successfully solving the tasks of vision, language, and vision-language~\citep{li2020unimo,li2021unimo,singh2021flava,DBLP:journals/corr/abs-2108-10904,wang2022ofa,diao2022prefix,wang2022omnivl}.

The state-of-the-art foundation models usually work the best for one type of tasks, namely language, vision, and vision-language. For example, RoBERTa~\citep{liu2019roberta}, BEiTv2~\citep{peng2022beit}, and X-VLM~\citep{zeng2021multi,zeng2022x} are language, vision, and vision-language foundation models respectively, and can achieve state-of-the-art performances for the specific type of tasks. It is still very challenging, however, to build a general foundation model that can perform the best in all types of tasks. Existing models, such as FLAVA~\citep{singh2021flava}, OFA~\citep{wang2022ofa}, DaVinci~\citep{diao2022prefix} and Uni-Perceiver-MoE~\citep{zhu2022uni}, are trying to achieve the goal. Their performances are still not satisfactory, however, when compared with the best performing foundation models for the individual types of tasks, as shown in Table~\ref{tab:compare_FMs}. Previous work~\cite {bingel2017identifying,wang2020makes} also shows that it is difficult to train a general foundation model in a multi-task learning setting that can effectively learn and utilize representations for all types of tasks. The reason is that language, vision, and vision-language are very different in nature, and a simple way of jointly training a model from language, vision, and vision-language data can easily create a suboptimal solution.


\begin{table*}[ht]
\centering
\resizebox{\textwidth}{!}{%
\begin{tabular}{l | c | c | c c c c c c}
\toprule
\multirow{3}*{Methods} & \multicolumn{1}{c}{Text Tasks} & \multicolumn{1}{c}{Vision Tasks} & \multicolumn{6}{c}{Multi-modal Tasks (MSCOCO Retriveal \& VQA \& NLVR)} \\ 
 & \multicolumn{1}{c}{GLUE} & \multicolumn{1}{c}{ImageNet} & \multicolumn{2}{c}{Zero-Shot} &
 \multicolumn{2}{c}{Fine-Tune} & \\ 
 \cmidrule(lr){2-3}
 \cmidrule(lr){4-4}
 \cmidrule(lr){5-9}
& MNLI & FT/LE & TR & IR & TR & IR & VQA & NLVR \\
\midrule
\multicolumn{9}{l}{\textit{Foundation models specifically for language, vision, or vision-language understanding}} \\
RoBERTa~\citep{liu2019roberta} & 87.6 & -- & -- & -- & -- & -- & -- & --\\
BEiTv2~\citep{peng2022beit} & -- & {\underline {85.5}}/80.1 & -- & -- & -- & -- & -- & -- \\
X-VLM~\citep{zeng2021multi} & -- & -- & 70.8/92.1/96.5 & 
55.6/82.7/90.0 & 80.4/95.5/98.2 & 63.1/85.7/91.6 & 78.1 & 84.8 \\
{\baby}~\citep{zeng2022x} & -- & -- & -- & 
-- & 83.5/96.3/{\underline {98.5}} & 66.2/87.1/92.2 & 80.4 & 87.0 \\
\midrule
\multicolumn{9}{l}{\textit{General foundation models}} \\
UNIMO-2~\citep{li2021unimo} & 87.5 & 80.8/- & -- & -- & -- & -- & 76.3 & -- \\
SimVLM~\citep{wang2021simvlm} & 83.4 & -/80.6 & -- & -- & -- & -- & 77.9 & 81.8 \\
FLAVA~\citep{singh2021flava} & 80.3 & -/75.5 & 42.7/76.8/- & 38.4/67.5/- & 61.5/82.1/89.6 & 50.1/74.4/83.2 & 72.8 & -- \\
OFA~\cite{wang2022ofa} & 84.3 & 82.2/-- & -- & -- & -- & -- & 78.0 & -- \\
DaVinci~\citep{diao2022prefix} & 83.1 & 83.9/78.8 & -- & -- & -- & -- & 76.3 & 77.9 \\
OmniVL~\citep{wang2022omnivl} & -- & -- & -- & -- & 76.8/93.6/97.3 & 58.5/82.6/89.5 & 78.3 & -- \\
Uni-Perceiver-MoE~\citep{zhu2022uni} & 81.5  & 84.5/-- & 64.6/--/-- & 51.6/--/-- & 70.5/--/-- & 54.1/--/-- & -- & -- \\
mPLUG-2$_{base}$~\citep{xu2023mplug} & 87.6 & --/-- & --/--/-- & --/--/-- & 81.2/95.2/98.1 & 65.3/86.9/92.4 & 79.3 & -- \\
{\bf \ModelNameB} & {\bf \underline {87.7}} & {\bf \underline{85.5}}/{\bf \underline {81.2}} & {\bf \underline {77.6/94.8/97.7}} & {\bf \underline {61.1/84.5/90.6}} & {\bf \underline {84.2/96.4}/98.4} & {\bf \underline {67.0/87.2/92.4}} & {\bf \underline {80.5}} & {\bf \underline {88.4}} \\
\bottomrule
\end{tabular}
}
\caption{\textbf{Performance comparisons between foundation models.}
All results are from \textit{base}-size models. MSCOCO is a cross-modal retrieval task, and IR and TR are image-retrieval and text-retrieval, respectively.
MNLI results are average accuracies of MNLI-m and MNLI-mm.
For ImageNet1k classification, we report linear evaluation (LE) performance and fine-tuning (FT) performance, respectively.
We report R@1/R@5/R@10 for all retrieval tasks at both zero-shot and fine-tune settings.
We report the VQA test-dev result and the NLVR test-P result.
\textbf{bold} denotes the best number across general foundation models.
\textbf{underline} denotes the best across all models. 
}
\label{tab:compare_FMs}
\end{table*}

To address the challenge, we propose a new method for training general foundation model, and bring in {\ModelName} (X-Foundation Model). {\ModelName} consists of three modular encoders for language (text) encoding, vision (image) encoding, and fusion encoding, as shown in Fig~\ref{Fig:model}. The language encoder, the vision encoder, and the entire model can be used in downstream tasks of language, vision, and vision-language understanding, respectively. The language encoder and the vision encoder follow the implementations of BERT~\citep{devlin2018bert} and ViT~\cite{dosovitskiy2020image}, respectively. Note that {\ModelName} do not include any extra parameters for language and vision tasks. The fusion encoder has the same architecture as BERT except that there is a cross-attention sub-layer after the self-attention sub-layer in each Transformer layer.

In learning of {\ModelName}, the language encoder, vision encoder, and fusion encoder are jointly trained with text data, image data, and image-text pair data as input. Given the text data, we train the language encoder by masked language modeling (MLM). Given the image data, we train the vision encoder by masked image modeling (MIM). Given the image-text pair data, we train the fusion encoder by image text matching (ITM), image-conditioned masked language modeling (IMLM), bounding box prediction (BBP), also train the vision encoder and the language encoder by image-text contrastive learning (ITC). (See Fig~\ref{Fig:model}.)


The essential thinking of our learning method is that language is more abstract than vision, and there is an asymmetric relationship between language and vision. Therefore, we separate the learning of the three encoders. The language encoder is trained mainly from text data and is isolated from the training of the fusion encoder. The vision encoder is simultaneously trained from image data and image-text pair data, guided by the vision-language training. The fusion encoder is trained from image-text pair data.

Our learning method includes two new techniques. One technique is to stop gradients from the vision-language training when learning the language encoder. The gradient flow is stopped from the fusion encoder to the language encoder in training, while the activation flow from the language encoder to the fusion encoder is as usual. As a result, the language encoder is not affected by training of the fusion encoder with image-text pair data. Moreover, the training of the fusion encoder concentrates on learning the alignments between language and vision features.

The other technique is to leverage the vision-language training to guide the learning of the vision encoder with masked image modeling (MIM). In MIM, the masked image is compared with the original image by the differences between the predicted representations and target representations at the masked and \texttt{[CLS]} positions. The vision encoder creates both the predicated and target representations, while there is gradient flow from the predicted representations but no gradient flow from the target representations. The vision encoder can create the target representations because it is also trained in the vision-language training.



We conduct experiments on a variety of twenty-three tasks of language, vision, and vision-language understanding. {\ModelName} can outperform other general foundation models by a large margin and can even achieve better or comparable performance than SOTA foundation models specifically designed for language, vision, or vision-language understanding tasks, as shown in Table~\ref{tab:compare_FMs}.

Our contribution is as follows.

(1) We address the problem of how to build a general foundation model that can perform the best for all the understanding tasks of language, vision, and vision-language.

(2) We propose a general foundation model, {\ModelName}, which can achieve better or competitive performance on both unimodal understanding tasks and multi-modal understanding tasks through two training techniques.

(3) The stop gradient technique is useful in maintaining text understanding capability and enhancing multi-modal understanding capability at the same time. We also propose a convenient method for mask image modeling with multi-modal learning. The technique can enhance both vision and multi-modal understanding.

\vspace{-0.3em}
\vspace{-0.3em}
\section{Related Work}

Following the success of language model pre-training~\citep{devlin2018bert,liu2019roberta,sun2019ernie,joshi2020spanbert,clark2020electra,lan2019albert,zhang2020ambert,he2020deberta}, vision pre-training~\citep{dosovitskiy2020image,he2022masked,bao2021beit,peng2022beit,wei2022masked} 
and vision-language pre-training~\citep{DBLP:conf/icml/RadfordKHRGASAM21,jia2021scaling,li2021align,li2022blip,yuan2021florence,wang2021vlmo,bao2022vl,zeng2021multi,zeng2022x} with Transformer as the backbone have also made significant progress recently, pushing the state-of-the-art of various understanding tasks of language, vision, and vision-language.

Recently, the fact that Transformer can model multi-modal data within a single architecture has inspired research to develop general foundation models that can solve language, vision, and vision-language tasks at the same time. 
UNIMO~\citep{li2020unimo,li2021unimo} jointly learns vision representations, language representations, and vision-language alignments in a shared space.
FLAVA~\citep{singh2021flava} performs pre-training with masked uni-modal and multi-modal modeling objectives. OFA~\citep{wang2022unifying} formulates vision-language tasks as sequence-to-sequence (seq2seq) problems and pre-trains a seq2seq model in multi-task learning. SimVLM~\citep{wang2021simvlm} pre-trains a seq2seq model with a single objective of language generation (prefix language modeling). 
DaVinci~\citep{diao2022prefix} combines prefix language modeling and prefix image modeling to learn a general foundation model for a wide range of tasks. 
Uni-Perceiver~\citep{zhu2021uni,zhu2022uni} builds a unified perception architecture that processes various modalities and tasks with a single Transformer and shared parameters.

Previous studies on general foundation models have shown that different capabilities can be established with only one model. Still, few studies demonstrate that the best performance can be achieved in all tasks with one model. In this paper, we propose a new method for training general foundation model and show that it can perform the best for all the understanding tasks of language, vision, and vision-language. We compare our model extensively with recent general foundation models on multiple dimensions, as shown in Appendix~\ref{appendix:model_comparisions}.

Several super-large foundation models (over 1B parameters) are proposed recently, most of which are trained on super-large {\it in-house} datasets (over 900M image-text pairs). The authors do not report results at the base (about 300M parameters) scale on {\it public} datasets, which we consider in this paper. CoCa~\citep{yu2022coca} pre-trains an image-text sequence-to-sequence model with contrastive loss and captioning loss. BEiT-3~\citep{wang2022image} uses a multi-way Transformer and a unified objective of masked ``language'' modeling for learning from image, text, and image-text pair data. 
Flamingo~\citep{alayrac2022flamingo} makes use of a large language model in vision-language pre-training to solve the ``in-context learning'' problem for vision-language tasks. PaLI~\citep{chen2022pali} jointly scales up the vision encoder and language encoder to cover a variety of language, vision, vision-language, and multilingual tasks.

\vspace{-0.3em}
\vspace{-0.3em}
\section{Method}
\label{sec:model_description}

\begin{figure*}[ht]
\begin{center}
\centerline{\includegraphics[width=0.9\textwidth]{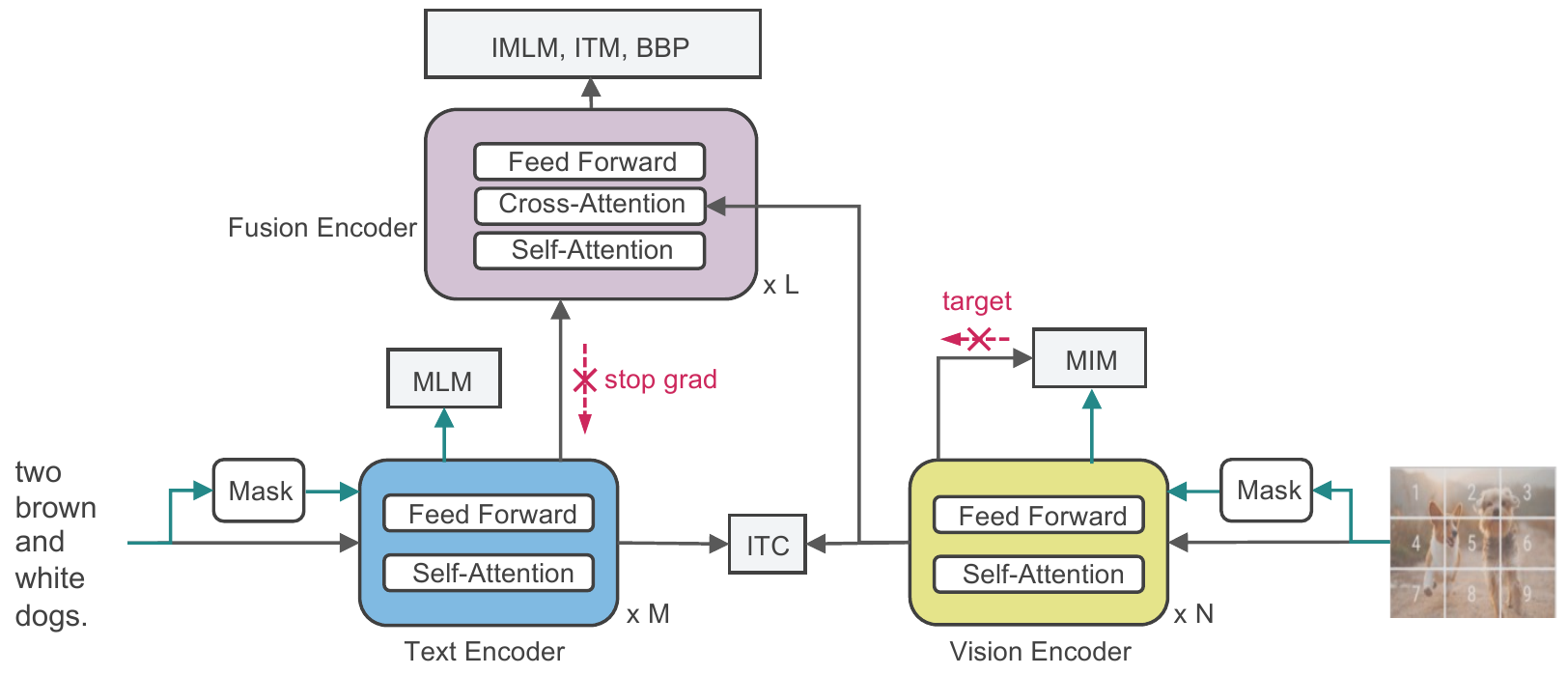}}
\caption{{\bf The architecture and pre-training process of {\ModelName}, a Transformer-based general foundation model.} 
Given a text, we learn the language encoder by MLM.
Given an image, we learn the vision encoder by MIM.
Given an image-text pair, we learn the fusion encoder by BBP, ITM, IMLM and ITC, and further learn the vision encoder by MIM. The gradients of BBP, ITM, and IMLM are stopped from the fusion encoder to the language encoder. The vision encoder is trained by MIM with both the image-text pair data and the image data. M, N and L denote numbers of encoder layers.}
\label{Fig:model}
\end{center}
\end{figure*}

\subsection{Model Architecture and Training Process}
\label{sec:model_arc}
We propose a new method for training general foundation model and bring in {\ModelName}, having a language encoder, a vision encoder,  and a fusion encoder, shown as Fig~\ref{Fig:model}. The architectures of language encoder, vision encoder and fusion encoder are following precious works~\citep{devlin2018bert,dosovitskiy2020image,li2021align}. We propose a new method for training general foundation model. Text, image, and image-text pair data are used as input to train {\ModelName}. The language encoder is trained by masked language modeling (MLM) and image text contrastive learning (ITC). The vision encoder is trained by masked image modeling (MIM) and ITC. The fusion encoder is trained by image text matching (ITM), image-conditioned masked language modeling (IMLM), and bounding box prediction (BBP). There are two new techniques developed for the training.
 

{\bf Stop Gradient.} 
We stop gradients from the vision-language training when learning the language encoder. Specifically, when the fusion encoder is trained with image-text pair data by ITM, IMLM, and BBP, there are forward flows (activations) from the language encoder to the fusion encoder, but there are no backward flows (gradients) from the fusion encoder to the language encoder. In this way, the language encoder is only trained with text data by MLM and with image-text pair data by ITC. The former helps the language encoder to learn text representations, and the latter helps to make alignments between text representations and image representations. Meanwhile, the training of the fusion encoder is performed separately with the focus of learning cross-modal alignments.

{\bf Vision-Language Guided Masked Image Modeling.} 
The training of vision encoder by MIM is carried out as follows. The image data is first masked and then predicted by the vision encoder. The differences between predicted representations and `target' representations at masked positions and \texttt{[CLS]} position are then measured with MSE (mean squared error) loss. The target representations are obtained from the same image data (without masking) by the vision encoder. There are no gradients from the target representations in the learning of the vision encoder. The vision encoder can create target representations because it is also trained with image-text pair data. In this way, the vision encoder is trained by both the cross-modal objectives (ITC, ITM, BBP, IMLM) with image-text pair data and the uni-modal objective (MIM) with image data. The representations obtained from the vision-language training are highly semantic, which is necessary for MIM as demonstrated in previous work~\citep{bao2021beit,peng2022beit,wei2022masked,wei2022mvp}.

There are mainly two advantages by exploiting the new MIM technique. First, it is convenient to conduct MIM with the signals from the vision-language training. Note that most previous work for MIM uses an external image tokenizer such as VQ-VAE~\citep{bao2021beit,singh2021flava}, CLIP~\citep{wei2022mvp}, and VQ-KL~\citep{peng2022beit}.
Second, the learning of the vision encoder and that of the fusion encoder are mutually enhanced. Once the vision encoder is trained, it is also utilized to train the fusion encoder. Fortunately, image data for training the vision encoder is relatively easy to obtain. 

\subsection{Pre-training Objectives}
\label{sec:objectives}

We explain six objectives in learning of {\ModelName}. Here, $\mathcal{T}$ represents the distribution of text data, $\mathcal{I}$ represents the distribution of image data, and $\mathcal{D}$ represents the distribution of image-text pair data.

{\bf Masked Language Modeling (MLM)}
We perform MLM on text data to learn the language encoder of {\ModelName}. Specifically we recover the masked tokens in a text by minimizing the cross entropy loss below.
{\setlength\abovedisplayskip{-0.2cm}
\setlength\belowdisplayskip{0.2cm}
\begin{equation}
\setlength{\abovedisplayskip}{0.2cm}
\setlength{\belowdisplayskip}{0.2cm}
\label{eqn:mlm}
\mathcal{L}_\mathrm{mlm} = \mathbb{E}_{T \sim \mathcal{T}} \mathrm{H} (\Vec{y}(\bar{T}), \hat{\Vec{p}}(\bar{T}{}))
\end{equation}}

where $T$ denotes a text, $\bar{T}$ denotes the masked text of $T$, $\hat{\Vec{p}}$ denotes the predicted probability vectors of masked tokens of $\bar{T}$, $\Vec{y}$ denotes the one-hot vectors representing the original tokens of $\bar{T}$, and $\mathrm{H}$ denotes cross-entropy. 

{\bf Image-Text Contrastive Learning (ITC).}
We use a contrastive loss as in CLIP~\citep{DBLP:conf/icml/RadfordKHRGASAM21} to learn the alignments between images and texts in ITC. Given a batch of images and texts, we calculate the cosine similarities between all image-text pairs. For each image, there is one text matched and the rest is unmatched. For each text, there is one image matched and the rest is unmatched. The contrastive loss is defined as follows.
{\setlength\abovedisplayskip{-0.2cm}
\setlength\belowdisplayskip{0.2cm}
\begin{align}
\label{eqn:itc}
\mathcal{L}_\mathrm{itc} = \frac{1}{2} \mathbb{E}_{(I,T)\sim \mathcal{D}} \big[ & \mathrm{H}(\Vec{y}^\mathrm{i2t}(I),\Vec{p}^\mathrm{i2t}(I)) \notag \\
& + \mathrm{H}(\Vec{y}^\mathrm{t2i}(T),\Vec{p}^\mathrm{t2i}(T)) \big]
\end{align}}
where $(I,T)$ denotes an image-text pair, $\Vec{p}^{\mathrm{i2t}}(I)$ denotes the in-batch image-to-text similarities, $\Vec{p}^{\mathrm{t2i}}(T)$ denotes the in-batch text-to-image similarities, $\Vec{y}^{\mathrm{i2t}}(I)$ denotes the one-hot vectors representing the image-to-text matching relations, $\Vec{y}^{\mathrm{t2i}}(T)$ denotes the one-hot vectors representing the text-to-image matching relations, and $\mathrm{H}$ is cross-entropy.

{\bf Image-Text Matching (ITM).}
We also learn the alignments between images and texts in ITM, using a loss indicating whether an image-text pair is matched. 
For each image in a batch there is a matched (positive) text, and we sample an unmatched (negative) text in the batch. For each text there is a matched (positive) image, and we sample an unmatched image in the batch.
The loss is defined as follows.

{\setlength\abovedisplayskip{-0.2cm}
\setlength\belowdisplayskip{0.2cm}
\begin{align}
\label{eqn:itm}
\mathcal{L}_\mathrm{itm} = \mathbb{E}_{(I,T)\sim \mathcal{D}} \big[ \mathrm{H} ({p}^\textrm{match}(I,T)) \notag \\
+ \mathrm{H} ({p}^\textrm{match}(\tilde{I},T)) \\
+ \mathrm{H} ({p}^\textrm{match}(I,\tilde{T})) \big] \notag
\end{align}}
where $(I,T)$ denotes a positive image-text pair, $(\tilde{I},T)$ and $(I,\tilde{T})$ denote negative image-text pairs, ${p}^{\mathrm{match}}(I,T)$ denotes a predicted matching probability of $(I,T)$, and $\mathrm{H}$ denotes logistic loss.

{\bf Image-conditioned Masked Language Modeling (IMLM)}
We conduct IMLM on image-text pair data to learn the fusion encoder. We recover the masked text tokens given for an image-text pair by minimizing the cross entropy loss below.

{\setlength\abovedisplayskip{-0.2cm}
\setlength\belowdisplayskip{0.2cm}
\begin{equation}
\label{eqn:imlm}
\mathcal{L}_\mathrm{imlm} = \mathbb{E}_{(I,T)\sim D} \mathrm{H} (\Vec{y}(\bar{T}), \hat{\Vec{p}}(I,\bar{T}))
\end{equation}}
where $(I,T)$ denotes an image-text pair, $\bar{T}$ denotes the masked text of $T$, $\hat{\Vec{p}}(I,\bar{T})$ denotes the predicted probability vectors of the masked tokens of $\bar{T}$ based on $I$, $\Vec{y}$ denotes the one-hot vectors representing the original tokens of $\bar{T}$, and $\mathrm{H}$ denotes cross-entropy. 

{\bf Bounding Box Prediction (BBP)}
We adopt the BBP in X-VLM~\citep{zeng2021multi,zeng2022x}, which locates the visual concept in the image by a bounding box given the text. With BBP we learn the alignments between the images and texts in multi-granularity. In BBP, two losses are simultaneously minimized to measure the differences between the predicted bounding box and the ground-truth bounding box. One is generalized intersection over union $GIoU$~\citep{rezatofighi2019generalized} and the other is $\ell_1$ distance.

{\setlength\abovedisplayskip{-0.2cm}
\setlength\belowdisplayskip{0.2cm}
\begin{equation}
\mathcal{L}_\mathrm{bbp} = \mathbb{E}_{(I, T)\sim D} \{ GIoU(\Vec{b}, \hat{\Vec{b}}) + \|\Vec{b}- \hat{\Vec{b}}\|_1 \}
\end{equation}}
where $\Vec{b}=(cx, cy, w, h)$ denotes the ground truth bounding box, $\hat{\Vec{b}}=(\hat{cx}, \hat{cy}, \hat{w}, \hat{h})$ denotes the predicted bounding box. A bounding box is represented by two coordinates, width, and height. 

\begin{table*}[t]
\centering
\resizebox{0.98\textwidth}{!}{
\begin{tabular}{lc|ccccccccccccccc}
\toprule
& & \multicolumn{1}{c}{RoBERTa} & \multicolumn{1}{c}{BEiTv2} &
\multicolumn{1}{c}{\babyx} &
\multicolumn{1}{c}{\babyx} &
\multicolumn{1}{c}{UNIMO-2} &
\multicolumn{1}{c}{FLAVA} & \multicolumn{1}{c}{SimVLM} & 
\multicolumn{1}{c}{OFA} & 
\multicolumn{1}{c}{DaVinci} & 
\multicolumn{1}{c}{DaVinci} &
\multicolumn{1}{c}{Uni-Per.} & 
\multicolumn{1}{c}{OmniVL} & 
\multicolumn{1}{c}{mPLUG-2$_{base}$} & 
\multicolumn{1}{c}{\ModelNameB} &
\multicolumn{1}{c}{\ModelNameB} \\
 &  & \small\texttt{1} & \small\texttt{2} &
\small\texttt{3} & \small\texttt{4} & \small\texttt{5} & \small\texttt{6} & \small\texttt{7} & \small\texttt{8} & \small\texttt{9} & \small\texttt{10} & \small\texttt{11} & \small\texttt{12} &
\small\texttt{13} & \small\texttt{14} & \small\texttt{15} \\
Task & Eval. & -- & -- & 4M & 1.3B & 4M & 70M & 1.8B & 21M & 46M & 648M & 30M & 14M & 17M & 4M & 1.3B \\

\midrule 
MNLI & FT & 87.6 & -- & -- & -- & 87.5 & 80.3 & 83.4 & 84.3 & 82.3 & 83.1 & 81.5 & -- & 87.6 & {\bf 87.7} & {\bf 87.7} \\
CoLA & FT & 63.6 & -- & -- & -- & 62.1 & 50.7 & 46.7 & 52.3 & 52.1 & 54.8 & 52.2 & -- & -- & 65.3 & {\bf 65.7} \\
MRPC & FT & 90.2 & -- & -- & -- & -- & 84.2 & 79.8 & 88.7 & 83.1 & 84.5 & -- & -- & 87.3 & {\bf 91.7} & 91.2 \\
QQP & FT & {\bf 91.9} & -- & -- & -- & -- & 88.7 & 90.4 & 91.3 & 88.2 & 88.9 & -- & -- & 91.3 &  91.8 & 91.7 \\
SST-2 & FT & 94.8 & -- & -- & -- & 94.7 & 90.9 & 90.9 & 92.7 & 90.5 & 91.4 & 90.9 & -- & 93.5 & {\bf 95.0} & 94.6 \\
QNLI & FT & 92.8 & -- & -- & -- & -- & 87.3 & 88.6 & 91.1 & 87.2 & 87.9 & 88.2 & -- & {\bf 93.2} & 92.9 & 92.8 \\
RTE & FT & 78.70 & -- & -- & -- & -- & 57.8 & 63.9 & 70.8 & 60.7 & 64.2 & 75.8 & -- & {\bf 85.2} & 83.8 & 82.7 \\
STS-B & FT & {\bf 91.2} & -- & -- & -- & {\bf 91.2} & 85.7 & 87.2 & -- & 86.3 & 87.1 & -- & -- & -- & 90.8 & 90.7 \\
\midrule 
\textbf{Language Avg.} &  & 86.4 & -- & -- & -- & -- & 78.2 & 78.9 & -- & 78.8 & 80.2 & -- & -- & -- & {\bf 87.4} & 87.1 \\
\midrule 
ImageNet & FT & -- & {\bf 85.5} & -- & -- & 80.8  & -- & -- & 82.2 & -- & 83.9 & 84.5 & -- & -- & 85.3 & {\bf 85.5} \\
ImageNet & LE & -- & 80.1 & -- &  -- & --  & 75.5 & 80.6 & 71.4$^\dagger$ & 75.9 & 77.7 & -- & -- & -- & 81.0 & {\bf 81.2} \\
Food101 & LE & -- & 88.2$^\dagger$ & -- & -- & -- & 88.5 & -- & 75.2$^\dagger$ & 89.3 & 90.1 & -- & 87.4  & -- & 88.7 & {\bf 90.5}  \\
CIFAR10 & LE & -- & 95.3$^\dagger$ & -- & -- & -- & 92.9 & -- & 86.1$^\dagger$ & 93.0 & 94.0 & -- & 96.2 & -- & 97.2 & {\bf 97.4} \\
CIFAR100 & LE & -- & 81.5$^\dagger$ & -- & -- & -- & 77.7 & -- & 66.7$^\dagger$ & 79.0 & 80.1 & -- & 83.2 & -- & {\bf 86.7} & 86.2 \\
Pets & LE & -- & {\bf 93.1$^\dagger$} & -- & -- & -- & 84.8 & -- & 81.0$^\dagger$ & 85.5 & 88.2 & -- & 87.1 & -- & 90.8 & 90.2 \\
DTD & LE & -- & {\bf 78.4$^\dagger$} & -- & -- & -- & 77.3 & -- & 70.3$^\dagger$ & 77.1 & 78.3 & -- & 76.2 & -- & 78.4 & {\bf 80.0} \\
Flowers102 & LE & -- & 95.7$^\dagger$ & -- & -- & -- & 96.4 & -- & 86.3$^\dagger$ & 96.1 & 96.9 & -- & 89.8 & -- & {\bf 97.1} & 96.4 \\
\midrule 
\textbf{Vision Avg.} & & -- & 88.7 & -- & -- & -- & 86.3 & -- & 79.2 & 86.7 & 87.9 & -- & 86.7 & -- & 89.8 & {\bf 90.1} \\
\midrule
VQAv2 & FT & -- & -- & 79.2 & 80.4 & 76.3 & 72.5 & 77.9 & 78.0 & 73.9 & 76.4 & -- & 78.3 & 79.3 & 79.1 & {\bf 80.5}  \\
NLVR2 & FT & -- & -- & 86.1 & 87.0 & -- & -- & 81.8 & -- & 77.9 & -- & -- & -- & -- & 86.7 & {\bf 88.4}  \\
Flickr30K TR R@1 & ZS & -- & -- & 85.1$^\dagger$ & 85.1$^\dagger$ & 84.6$^\dagger$ & 88.5 & 67.7 &-- & -- & -- & 82.1 & -- & -- & 90.1 & {\bf 93.4}  \\
Flickr30K IR R@1 & ZS & -- & -- & 77.3$^\dagger$ & 79.2$^\dagger$ & 72.7 & 65.2 & -- & -- & -- & -- & 72.4 & -- & -- & 79.1 & {\bf 84.1} \\
Flickr30K TR R@1 & FT & -- & -- & 97.4 & {\bf 98.5} & 92.0 & -- & -- & -- & -- & -- & 93.6 & 94.9 & 96.9 & 97.4 & 98.1 \\
Flickr30K IR R@1 & FT & -- & -- & 90.0 & {\bf 90.4} & 80.1 & -- & -- & -- & -- & -- & 79.8 & 83.4 & 88.2 & 88.6 & 89.9 \\
COCO TR R@1 & ZS & -- & -- & 68.4$^\dagger$ & 71.7$^\dagger$ & -- & 42.7 & -- & -- & -- & -- & 64.6 & -- & -- & 73.8 & {\bf 77.6} \\
COCO IR R@1 & ZS & -- & -- & 55.2$^\dagger$ & 58.3$^\dagger$ & -- & 38.4 & -- & -- & -- &  -- & 51.6 & -- & -- & 59.4 & {\bf 61.1} \\
COCO TR R@1 & FT & -- & -- & 80.5 & 83.5 & -- & -- & -- & -- & -- & -- & 70.5 & 76.8  & 81.2 &  81.8 & {\bf 84.2} \\
COCO IR R@1 & FT & -- & -- & 62.7 & 66.2 & -- & -- & -- & -- & -- & -- & 52.6 & 58.5 & 65.3 & 64.7 & {\bf 67.0} \\
\midrule
\multicolumn{2}{l}{\textbf{Vision-Language Avg.}} & -- & -- & 78.2 & 80.0 & -- & -- & -- & -- & -- & -- & -- & -- & -- & 80.1 & {\bf 82.4} \\
\bottomrule
\end{tabular}}
\caption{
\textbf{Experimental results on vision, language and vision-language tasks.}
The multi-modal data size used for pre-training are reported under the model name.
MNLI results are average of MNLI-m and MNLI-mm. 
MRPC results are average accuracies and F1 scores.
Matthews correlation coefficient (MCC) is reported for CoLA, and Pearson correlation coefficient (PCC) is reported for STS-B.
We report accuracies for all the vision and multi-modal tasks. FT is short for fine-tuning, LE for linear evaluation, ZS for zero-shot, TR for text retrieval, and IR for image retrieval. Results for RoBERTa are from its corresponding paper~\citep{liu2019roberta}, and they use the mid-training~\citep{Phang2018SentenceEO} on MNLI for RTE, MRPC, and STS-B while other models (e.g., DaVinci, {\ModelName}) do not use this trick. Note that mPLUG-2 used more layers and parameters than RoBERTa and {\ModelName} for the language understanding tasks.  
Language Avg. is the average score of all the language tasks, while Vision Avg. is the average score of six line evaluation tasks except ImageNet.
Vision-Language Avg. is the average score of all vision-language tasks.
$^\dagger$ are our reproduced results with the officially released models.}
\label{tab:main_result}
\end{table*}

{\bf Masked Image Modeling (MIM)}
We perform MIM on image data and image-text pair data to learn the vision encoder. Specifically, we recover the masked image patches in an image by minimizing the loss below.

{\setlength\abovedisplayskip{-0.2cm}
\setlength\belowdisplayskip{0.2cm}
\begin{align}
\label{eqn:mim}
\mathcal{L}_\mathrm{mim} = \mathbb{E}_{(I, T)\sim \mathcal{D}}||\Vec{v}(\bar{I})- \hat{\Vec{v}}(\bar{I})||_2 \notag \\ 
+ \mathbb{E}_{I\sim \mathcal{I}}||\Vec{v}(\bar{I}) - \hat{\Vec{v}}(\bar{I})||_2 
\end{align}}

where $(I, T)$ and $I$ denote an image-text pair and a single image respectively, $\bar{I}$ denotes the masked image $I$, $\hat{\Vec{v}}(\bar{I})$ denotes the predicted representations at the masked positions and \texttt{[CLS]} of $\bar{I}$, and ${\Vec{v}}(\bar{I})$ denotes the target representations at the masked positions and \texttt{[CLS]} of $\bar{I}$. 
$||\dot||_2$ is the MSE loss. We employ block masking following previous work~\citep{bao2021beit,peng2022beit}. Note that $(I, T)$ and $I$ are independently sampled from $\mathcal{D}$ and $\mathcal{I}$, and the sample sizes are not necessarily equal.

Finally, the pre-training objective of {\ModelName} is defined as the sum of the losses described above.
{
\begin{equation}
\mathcal{L} = \mathcal{L}_\mathrm{mlm} + \mathcal{L}_\mathrm{itc} + \mathcal{L}_\mathrm{itm} + \mathcal{L}_\mathrm{imlm} +  \mathcal{L}_\mathrm{bbp} + 
\mathcal{L}_\mathrm{mim} \nonumber
\end{equation}}
\vspace{-0.3em}
\vspace{-0.3em}
\section{Experiments}

\subsection{Implementation Details}
\label{sec:impl_details}

\paragraph{Pre-training Datasets.}

We mainly conduct our experiments on several widely used public datasets, 
consisting of two in-domain datasets, COCO~\cite{lin2014microsoft} and Visual Genome (VG)~\cite{krishna2016visual}, and two out-of-domain datasets, SBU Captions~\cite{ordonez2011im2text} and Conceptual Captions (CC)~\cite{sharma2018conceptual}. Following X-VLM~\citep{zeng2021multi,zeng2022x}, we also include annotations of objects and regions from RefCOCO~\cite{yu2016modeling}, Objects365~\cite{shao2019objects365} and OpenImages~\cite{kuznetsova2018open}. Since we assume also using uni-modal data, we include RoBERTa corpus~\citep{liu2019roberta}, $C4$ datasets~\citep{raffel2020exploring} and Imagenet21K~\citep{ridnik2021imagenet21k}. In addition, we also scale up the pre-training dataset with Conceptual 12M dataset (CC-12M)~\cite{changpinyo2021conceptual} and LAION~\cite{schuhmann2022laion} as the ``more data" setting, which contains around 1.3B image-text pairs. Please refer to Appendix~\ref{appendix:details_of_datasets} for statistics of the pre-training datasets.


\begin{table*}[t]
\centering
\resizebox{\textwidth}{!}{
\begin{tabular}{lc|cccc|cccc}
\toprule
\multirow{3}{*}{\bf Model} & \multirow{3}{*}{\textbf{\# Params}} & \multicolumn{2}{c}{\bf MSCOCO (5K test set)} & \multicolumn{2}{c}{\bf Flickr30K (1K test set)} & \multicolumn{2}{c}{\bf MSCOCO (5K test set)} & \multicolumn{2}{c}{\bf Flickr30K (1K test set)} \\
 & & {\bf TR-Fine-Tune} & {\bf IR-Fine-Tune} & {\bf TR-Fine-Tune} & {\bf IR-Fine-Tune} & {\bf TR-Zero-Shot} & {\bf IR-Zero-Shot} & {\bf TR-Zero-Shot} & {\bf IR-Zero-Shot}  \\
 & & R@1/R@5/R@10 & R@1/R@5/R@10 & R@1/R@5/R@10 & R@1/R@5/R@10 & R@1/R@5/R@10 & R@1/R@5/R@10 & R@1/R@5/R@10 & R@1/R@5/R@10 \\
\midrule
ALBEF & 210M & 73.1/91.4/96.0 & 56.8/81.5/89.2 & 94.3/99.4/99.8 & 82.8/96.7/98.4 & -- & -- & {\bf 90.5}/98.8/99.7 & 76.8/93.7/96.7 \\
VLMo$_\mathrm{base}$ & 175M &  74.8/93.1/96.9 &  57.2/82.6/89.8 & 92.3/99.4/99.9 & 79.3/95.7/97.8 & -- & -- & -- & -- \\
VL-BEiT & 175M & 79.5/--/-- & 61.5/--/-- & 95.8/--/-- & 83.9/--/-- & -- & -- & -- & -- \\
OmniVL & 288M & 76.8/93.6/97.3 & 58.5/82.6/89.5 & 94.9/9.6/99.9 & 83.4/97.0/98.6 & -- & -- & -- & -- \\
X-VLM & 216M & 80.4/95.5/98.2 & 63.1/85.7/91.6 & 96.8/99.8/{\bf 100} & 86.1/97.4/98.7 & 70.8/92.1/96.5 & 55.6/82.7/90.0 & 85.3/97.8/99.6 & 71.9/93.3/96.4 \\
\babyx$_\mathrm{base}$ & 255M & 80.5/95.5/97.8 & 62.7/84.7/90.7 & {\bf 97.4}/99.9/{\bf 100} & {\bf 90.0/98.6/99.3} &
68.4$^\dagger$/92.5$^\dagger$/96.8$^\dagger$ & 55.2$^\dagger$/82.2$^\dagger$/89.3$^\dagger$ & 85.1$^\dagger$/{\bf 99.2$^\dagger$/100.0$^\dagger$} & 77.3$^\dagger${\bf /95.3$^\dagger$/97.6$^\dagger$} \\
{\bf \ModelNameB} & 327M & {\bf 81.8/96.0/98.3} & {\bf 64.7/86.1/91.6} & {\bf 97.4/100/100} & 88.6/97.9/98.9 & {\bf 73.8/93.9/97.2} & {\bf 59.4/83.6/90.0} & 90.1/{\bf 99.2}/99.9 & {\bf 79.1}/95.2/97.3 \\





\midrule
\multicolumn{10}{l}{\textit{More Data}} \\
CLIP & ~490M & -- & -- & 88.7/98.0/99.2 & 76.7/93.6/96.4 & 58.4/81.5/88.1 & 37.8/62.4/72.2 &
88.0/98.7/99.4 & 68.7/90.6/95.2 \\
ALIGN & 490M & 77.0/93.5/96.9 & 59.9/83.3/89.8 & 95.3/99.8/100 & 84.9/97.4/98.6 & 58.6/83.0/89.7 & 45.6/69.8/78.6 & 88.6/98.7/99.7 & 75.7/93.8/96.8 \\
Florence & 893M & 81.8/95.2/-- & 63.2/85.7/-- & 97.2/99.9/-- & 87.9/98.1/-- & 64.7/85.9/-- &
47.2/71.4/-- & 90.9/99.1/-- & 76.7/93.6/-- \\
\babyx$_\mathrm{base}$ & 255M & 83.5/96.3/{\bf 98.5} & 66.2/87.1/92.2 & {\bf 98.5/100/100} & {\bf 90.4}/98.2/99.3 &
71.7$^\dagger$/93.4$^\dagger$/97.5$^\dagger$ & 58.3$^\dagger$/{\bf 84.7$^\dagger$/91.0$^\dagger$} & 84.6$^\dagger$/99.1$^\dagger$/{\bf 99.9$^\dagger$} & 79.2$^\dagger$/96.4$^\dagger$/98.0$^\dagger$ \\
{\bf \ModelNameB} & 327M & {\bf 84.2/96.4}/98.4 & {\bf 67.0/87.2/92.4} & 98.1/{\bf 100/100} & 89.9/{\bf 98.6/99.4} & {\bf 77.6/94.8/97.7} & {\bf 61.1}/84.5/90.6 & {\bf 93.4/99.8/99.9} & {\bf 84.1/96.5/98.1} \\
\midrule
\midrule
\multicolumn{6}{l}{\textit{Super-Large Models}} \\



\textcolor{gray}{CoCa} & \textcolor{gray}{2.1B} & \textcolor{gray}{--} & \textcolor{gray}{--} & \textcolor{gray}{--} & \textcolor{gray}{--} &
\textcolor{gray}{66.3/86.2/91.8} &
\textcolor{gray}{51.2/74.2/82.0} &
\textcolor{gray}{92.5/99.5/99.9} &
\textcolor{gray}{80.4/95.7/97.7} \\
\textcolor{gray}{BEiT-3} & \textcolor{gray}{1.9B} & \textcolor{gray}{84.8/96.5/98.3} & \textcolor{gray}{67.2/87.7/92.8} & \textcolor{gray}{98.0/100/100} & \textcolor{gray}{90.3/98.7/99.5} &
\textcolor{gray}{--} & \textcolor{gray}{--} & \textcolor{gray}{94.9/99.9/100.0} & \textcolor{gray}{81.5/95.6/97.8} \\

\bottomrule
\end{tabular}}
\caption{Results of text-retrieval (TR) and image-retrieval (IR) on COCO and Flickr30K.
$^\dagger$ denotes our reproduced results with the officially released models. 
In more data setting, we use Conceptual 12M dataset (CC-12M)~\cite{changpinyo2021conceptual} and LAION~\cite{schuhmann2022laion} as additional datasets. More details are explained in Appendix~\ref{appendix:details_of_datasets}. 
Giant models with over 1B parameters (e.g., BEiT-3) are in grey since they are not directly comparable with other models.
}
\label{tab:retrieval}
\end{table*}

\begin{table*}[t]
\centering
\resizebox{0.98\textwidth}{!}{
\begin{tabular}{lc|cccccccccccccccccc}
\toprule
& & \rotatebox{90}{ImageNet} & \rotatebox{90}{Food101} & \rotatebox{90}{CIFAR10} & \rotatebox{90}{CIFAR100} & \rotatebox{90}{StanfordCars} & \rotatebox{90}{Aircraft} & \rotatebox{90}{DTD} & \rotatebox{90}{OxfordiiitPets} & \rotatebox{90}{Flowers102} & \rotatebox{90}{MNIST} & \rotatebox{90}{STL10} & \rotatebox{90}{Country211} & \rotatebox{90}{Sun397} & \rotatebox{90}{SST} & \rotatebox{90}{Caltech101} & \rotatebox{90}{GTSRB} & \rotatebox{90}{PCAM} & \rotatebox{90}{FER2013} \\
\midrule 
CLIP & B/16-224px & 80.2 & {\bf 92.8} & 96.2 & 83.1 & 86.7 &  {\bf 59.5} & 79.2 & {\bf 93.1} & {\bf 97.1} & {\bf 99.0} & 99.0 & {\bf 30.1} & 78.4 & {\bf 75.5} & 94.7 & 86.6 & 83.5 & 69.5 \\
FLAVA & B/16-224px & 75.5 & 88.5 & 92.9 & 77.7 & 70.9 & 47.3 & 77.3 & 84.8 & 98.1 & {\bf 99.0} & 98.9 & 28.9 & 82.1 & 57.1 & 95.7 & 79.5 & {\bf 85.3} & 61.1\\
DaVinci & B/16-224px & 77.7 & 90.1 & 94.0 & 80.1 & 74.6 & 49.6 & 78.3 & 88.2 & 96.9 & {\bf 99.0} & {\bf 99.2} & 29.9 & -- & -- & -- & -- & -- & --\\
{\bf \ModelNameB} & B/16-224px & {\bf 81.2} & 90.5 & {\bf 97.4} & {\bf 86.2} & {\bf 88.3} & 47.4 & {\bf 80.0} & 90.2 & 96.4 & {\bf 99.0} & {\bf 99.2} & 24.9 & {\bf 93.9} & 60.6 & {\bf 97.1} & {\bf 90.9} & 82.4 & {\bf 72.6} \\
\bottomrule
\end{tabular}}
\caption{
\textbf{Linear evaluation performance of four foundation models over 18 datasets.} B/16-224px means base size model, 16*16 patches, and 224*224 resolution, respectively. The best performance is identified with bold.
}
\label{tab:image_classifications}
\end{table*}

\paragraph{Pre-training Settings.}
Our model is of base size, and the detailed parameters are explained in Appendix~\ref{appendix:details_of_hyper_params}. The vision encoder is initialized with BEiTv2. The language encoder is initialized with RoBERTa. The fusion encoder is trained from scratch. {\ModelName} is pre-trained at image resolution of $224\times224$ with patch size of $16\times16$. 
We pre-train {\ModelName} for 200K steps with a batch size of 3072 image-text pairs, 3072 images, and 8192 sentences on 32 A100, 
which takes about six days. The learning rate for both models is warmed-up to $1e^{-4}$ in the first 2500 steps and decayed following a linear schedule. 
We set the maximum number of text tokens to 30 for image-text pairs, while that of pure text corpus is set to 128. For the ``more data" setting, we pre-train {\ModelName} for 400k steps with 18k batch size on 64 A100. Due to the consideration of computational cost, we did not pre-train the large or giant models. We apply mixed precision for pre-training. We choose widely used downstream tasks whose details are shown in Appendix~\ref{appendix:details_of_downstream_tasks}.


\subsection{Comparison with Foundation Models}
\label{sec:exp_fm}

We extensively compare the performance of {\ModelName} with state-of-the-art foundation models on vision, language, and multi-modal tasks. We first compare our model with general foundation models, including UNIMO-v2~\citep{li2021unimo}, FLAVA~\citep{singh2021flava}, SimVLM~\citep{wang2021simvlm}, OFA~\citep{wang2022ofa}, DaVinci~\citep{diao2022prefix}, 
Uni-Perceiver-MoE~\citep{zhu2022uni},
OmniVL~\citep{wang2022omnivl}, and mPLUG-2~\citep{xu2023mplug}. We also include comparisons with SOTA foundation models specifically designed for language, vision, or vision-language tasks, RoBERTa~\citep{liu2019roberta}, BEiTv2~\citep{peng2022beit}, and {\babyx}~\citep{zeng2022x}. 
There are several observations in Table~\ref{tab:main_result}.
First, {\ModelNameB} (column 15) outperforms all the previous general foundation models (column 5-13) across almost all tasks by a large margin, becoming a new and stronger general foundation model. When we use less pre-training data, {\ModelName} can also achieve competitive performance compared with previous general foundation models (column 5-13 vs 14).
Second, we compare {\ModelName} with state-of-the-art foundation models specifically designed for language, vision, and vision-language tasks, RoBERTa, BEiTv2 and {\babyx}. We observe that {\ModelName} is also better than or comparable with the foundation models (column 1,2,3,4 vs 15).
We further compare our model, {\ModelNameB}, with three previous foundation models on 18 image classification tasks on the linear evaluation setting to evaluate generalization performance on vision understanding tasks. The results are shown in Table~\ref{tab:image_classifications}. {\ModelNameB} wins 11 of 18 tasks, 7 for CLIP, 2 for FLAVA, and 2 for DaVinci.


\begin{table*}[t]
\centering
\resizebox{0.98\textwidth}{!}{
\begin{tabular}{lc|c|ccc|ccc|ccc}
\toprule
& & \multicolumn{10}{c}{\ModelNameB} \\
& & \multicolumn{1}{c}{RoBERTa$^\dagger$} & \multicolumn{1}{c}{S-MLM} &
\multicolumn{1}{c}{S-ITM} &
\multicolumn{1}{c}{wostop} &
\multicolumn{1}{c}{BEiTv2$^\dagger$} &
\multicolumn{1}{c}{woMIM} & 
\multicolumn{1}{c}{wBEiTv2 Tokenizer} &
\multicolumn{1}{c}{\babyx$^\dagger$} &
\multicolumn{1}{c}{Multi-task} &
\multicolumn{1}{c}{\bf ALL} \\
Task & Eval. & \small\texttt{1} & \small\texttt{2} &
\small\texttt{3} & \small\texttt{4} & \small\texttt{5} & \small\texttt{6} & \small\texttt{7} & \small\texttt{8} & 
\small\texttt{9} & \small\texttt{10}\\

\midrule 
MNLI & FT & {\bf 87.7} & 87.4 & 87.3 & {\bf 87.7} & -- & -- & -- & -- & 87.4 & 87.6 \\
CoLA & FT & 63.2 & 61.6 & 63.6 & 64.2  & -- & -- & -- & -- & 62.2 & {\bf 65.2} \\
MRPC & FT & 90.7 & 92.2 & 91.1 & 90.7  & -- & -- & -- & -- & 92.0 & {\bf 92.5} \\
QQP & FT & 91.5 & {\bf 91.6} & {\bf 91.6} & {\bf 91.6}  & -- & -- & -- & -- & {\bf 91.6} & {\bf 91.6} \\
SST-2 & FT & 95.0 & 95.1 & 94.2 & 94.6  & -- & -- & -- & -- & 94.4 & {\bf 95.3} \\
QNLI & FT & 93.1 & 93.0 & {\bf 93.2} & 92.5  & -- & -- & -- & -- & 92.8 & 92.9  \\
RTE & FT & 80.9 & 79.1 & 81.6 & 81.2  & -- & -- & -- & -- & 79.8 & {\bf 81.9} \\
STS-B & FT & {\bf 90.9} & 90.7 & 90.7 & 90.4  & -- & -- & -- & -- & 90.1 & 90.8 \\
\midrule 
\textbf{Language Avg.} &  & 86.6 & 86.4 & 86.7 & 86.6  & -- & -- & -- & -- & 86.3 & {\bf 87.2} \\
\midrule 
ImageNet & FT & -- & -- & -- & --  & {\bf 85.5} & 84.8 & 85.0 & -- & 85.0 & 85.3 \\
ImageNet & LE & -- & -- & -- &  --  & 80.5 & 79.1 & 79.4 & -- & 79.3 & {\bf 81.1} \\
Food101 & LE & -- & -- & -- & -- & 88.2 & 86.9 & 87.2 & -- & 86.9 & {\bf 88.7}\\
CIFAR10 & LE & -- & -- & -- & -- & 95.3 & 96.6 & 96.5 & -- & 96.6 & {\bf 97.5} \\
CIFAR100 & LE & -- & -- & -- & -- & 81.5 & 83.3 & 83.9 & -- & 84.1 & {\bf 86.9} \\
Pets & LE & -- & -- & -- & -- & {\bf 93.1} & 88.1 & 88.5 & -- & 88.2 & 90.7 \\
DTD & LE & -- & -- & -- & -- & 78.4 & 77.7 & 76.9 & -- & 78.0 & {\bf 78.7} \\
Flowers102 & LE & -- & -- & -- & -- & 95.7 & 94.1 & 94.5 & -- & 94.2 & {\bf 97.1}  \\
\midrule 
\textbf{Vision Avg.} & & -- & -- & -- & -- & 87.3 & 86.3 & 86.5 & -- & 86.5 & {\bf 88.2} \\
\midrule
VQAv2 & FT & -- & {\bf 78.8} & 78.5 & 78.7 & -- & 78.3 & 78.2 & 78.0 & 78.2 & 78.6 \\
NLVR2 & FT & -- & 86.3 & 86.0 & 86.4 & -- & 85.9 & 85.5 & 86.2 & 86.1 & {\bf 86.7} \\
Flickr30K TR R@1 & ZS & -- & 88.3 & 87.2 & 87.1 & -- & 87.1 & 87.2 & 87.7 & 85.0 &  {\bf 89.3} \\
Flickr30K IR R@1 & ZS & -- & 76.6 & 74.9 & 75.8 & -- & 76.1 & 75.3 & 75.1 & 75.6 & {\bf 77.4} \\
Flickr30K TR R@1 & FT & -- & 97.5 & 97.0 & 97.2 & -- & 96.4 & 96.7 & 97.0 & 97.0 & {\bf 97.7} \\
Flickr30K IR R@1 & FT & -- & {\bf 87.4} & 86.9 & 87.3 & -- & 86.2 & 86.6 & 86.2 & 86.4 & {\bf 87.4} \\
COCO TR R@1 & ZS & -- & 72.0 & 72.1 & 70.5 & -- & 73.0 & 72.1 & {\bf 73.2} & 69.9 & 72.8 \\
COCO IR R@1 & ZS & -- & 58.4 & 57.1 & 57.7 & -- & 58.2 & 57.7 & 57.7 & 56.5 & {\bf 59.0} \\
COCO TR R@1 & FT & -- & {\bf 81.2} & 80.2 & 80.9 & -- & 80.6 & 80.1 & 80.3 & 80.0 & {\bf 81.2} \\
COCO IR R@1 & FT & -- & {\bf 64.2} & 63.4 & 63.6 & -- & 63.7 & 63.0 & 63.1 & 63.0 & 64.0 \\
\midrule
\textbf{Vision-Language Avg.} &  & -- & 79.1 & 78.3 & 78.5 & -- & 78.6 & 78.2 & 78.5 & 77.8 & {\bf 79.4} \\
\bottomrule
\end{tabular}}
\caption{
\textbf{Ablation studies on vision, language, and vision-language tasks.} We use the same settings as Table~\ref{tab:main_result}. ``ALL'' means we use both of our proposed techniques.
To compare fairly, we pre-train all variants with the same data at the same settings for both pre-training and fine-tuning.
Avg. means the average score.
}
\label{tab:ablation_result}
\end{table*}

\subsection{Comparison with multi-modal Models}
\label{sec:exp_mm}
In addition to general foundation models, we also compare {\ModelName} with state-of-the-art vision-language models.
The results are shown in Table~\ref{tab:retrieval} and Table~\ref{tab:results_mm}.
{\ModelName} demonstrates its superiority on five downstream vision-language tasks including MSCOCO Retrieval, Flick Retrieval, VQA, NLVR and RefCOCO+. Note that {\ModelNameB} outperforms CLIP, ALIGN and Florence on image-text retrieval tasks with fewer parameters and much less training data. Compared to the recently released SOTA vision-language model, {\babyx}, {\ModelName} is much better on zero-shot image-text retrieval tasks. When we scale up pre-training datasets, {\ModelNameB} is consistently better than previous vision-language models for most cases.


\begin{table}[!t]
\centering	
\resizebox{0.98\columnwidth}{!}{
	\begin{tabular}	{lccccccccc} 
	\toprule
	 \multirow{2}{*}{Method} & \multirow{2}{*}{\# Params} & \multicolumn{2}{c}{VQA} & \multicolumn{2}{c}{NLVR2} & \multicolumn{3}{c}{RefCOCO+} \\
	  & & test-dev & test-std & dev & test-P & val & testA$^d$ & testB$^d$ \\
\midrule
ALBEF & 210M & 74.5 & 74.7 & 80.2 & 80.5 & -- & -- & --  \\
VLMo$_\mathrm{base}$ & 175M & 76.6 & 76.9 & 82.8 & 83.3 & -- & -- & --  \\
METER & 341M & 77.7 & 77.6 & 82.3 & 83.1 & -- & -- & --  \\
VL-BEiT & 175M & 77.5 & 77.8 & 81.9 & 82.7 & -- & -- & --  \\
BLIP$_\mathrm{base}$ & 240M & 78.2 & 78.2 & 82.5 & 83.1 & -- & -- & --  \\
X-VLM & 216M & 78.1 & 78.1 & 84.2 & 84.2 & 80.2 & 86.4 & 71.0 \\
OFA$_\mathrm{base}$ & 182M & 78.0 & 78.1 & -- & -- & 81.4 & 87.2 & 74.3  \\
OmniVL & 288M & 78.3 & 78.4 & -- & -- & -- & -- & -- \\
\babyx$_\mathrm{base}$ & 255M & {\bf 79.2} & {\bf 79.3} & 85.9 & 86.1 & {\bf 85.4} & 89.2 & 77.3 \\
{\bf \ModelNameB} & 327M & 79.1 & 79.2 & {\bf 86.3} & {\bf 86.5} & 84.8 & {\bf 89.7} & {\bf 79.1} \\

\midrule

\multicolumn{6}{l}{\textit{More Data}} \\
SimVLM$_\mathrm{base}$ & 273M & 77.9 & 78.1 & 81.7 & 81.8 & -- & -- & -- \\
\babyx$_\mathrm{base}$ & 255M & 80.4 & 80.2 & 86.2 & 87.0 & 85.2 & 90.3 & 78.4 \\
{\bf \ModelNameB} & 327M & {\bf 80.5} & {\bf 80.4} & {\bf 87.6} & {\bf 88.4} & {\bf 86.1} & {\bf 90.4} & {\bf 79.8} \\

\midrule
\midrule
\multicolumn{6}{l}{\textit{Super-Large Models}} \\
\textcolor{gray}{CoCa} & \textcolor{gray}{2.1B} & \textcolor{gray}{82.3} & \textcolor{gray}{82.3} & \textcolor{gray}{86.1} & \textcolor{gray}{87.0} & \textcolor{gray}{--} & \textcolor{gray}{--} & \textcolor{gray}{--} \\
\textcolor{gray}{BEiT-3} & \textcolor{gray}{1.9B} & \textcolor{gray}{84.2} & \textcolor{gray}{84.0} & \textcolor{gray}{91.5} & \textcolor{gray}{92.6} & \textcolor{gray}{--} & \textcolor{gray}{--} & \textcolor{gray}{--} \\

\bottomrule  	  
\end{tabular}}
\caption
{Results on VQA, visual reasoning and visual grounding. Giant models with over 1B parameters (e.g., CoCa and BEiT-3) are in grey because they are not directly comparable with other models.}
\label{tab:results_mm}
\end{table}


\subsection{Ablation Study}
\label{sec:exp_ablation}

To verify the contributions of different modules in our framework, we ablate them and evaluate the performance of {\ModelName} on all downstream tasks. The results are shown in Table~\ref{tab:ablation_result}.
We first explain several abbreviations in the table. S-MLM means that we only stop the gradient of language representations in IMLM task, while S-ITM means stopping the gradient of language representations for computing ITM and BBP. wostop indicates without stopping the gradients of all language representations. woMIM means that we do not learn by MIM, while wBEiTv2 tokenizer means that we learn by MIM with the image tokenizer used in BEiTv2. Multi-task is a variation that uses straightforward multi-task learning to optimize the three encoders in {\ModelName}. To make a fair comparison, we also train RoBERTa, BEiTv2 and {\babyx} with the same data noted as RoBERTa$^\dagger$, BEiTv2$^\dagger$ and {\babyx$^\dagger$}. Note that we also increase the fusion layers in {\babyx$^\dagger$} to make the parameter sizes comparable to our models. RoBERTa$^\dagger$, BEiTv2$^\dagger$ and {\babyx$^\dagger$} all have slightly better results on average than the official ones.
From the results, we have the following observations.

First, both designs (stop gradient and vision-language guided MIM) bring improvements, and the combination can make further improvements on all three downstream tasks (column 10 vs. others).
Second, without separated language representations, models always perform worse on language understanding tasks (column 10 vs. 2,3,4). Besides, the separate language representations in the IMLM task on image-text data are helpful for multi-modal tasks (column 2 vs. 4). As we point out in section~\ref{sec:intro}, the fusion encoder can concentrate on learning the alignments between language and vision features instead of predicting masked tokens with clues from other visible text tokens. Although S-ITM shows slight side effects (column 4 vs. 3), stopping the gradients of language representation in the fusion encoder is necessary to simultaneously achieve strong language understanding and vision-language understanding capability.
Third, the vision-language guided MIM task is useful for both vision-language and vision learning (column 10 vs. 6). Meanwhile, the targets in our MIM task are better than the BEiTv2 tokenizer (column 10 vs. 7).
Four, {\ModelName} is much better than a naive multi-task learning strategy for a foundation model, compared with which, {\ModelNameB} improves an average of 0.9\%, 1.7\% and 1.6\% on language, vision, and vision-language tasks, respectively (column 10 vs. 9).
Five, {\ModelName} is also better than foundation models specifically designed for language, vision, and vision-language tasks with the same training corpus (column 10 vs. 1,5,8).
\vspace{-0.3em}
\vspace{-0.3em}
\section{Limitations and Potential Risks}
\label{appendix:limitations}

\paragraph{Limitations.}
Like most existing work on foundation models, the entire project consumed over 5 A100 GPU years on a computing cluster with high electricity costs, although we only tested base models. There is still potential for efficiency improvement through sparse attention~\citep{zaheer2020big} or the lottery ticket hypothesis~\citep{frankle2018lottery}. We will explore the techniques to improve the training efficiency and reduce the carbon footprint so that we can adhere to the proposals on ``green'' deep learning~\citep{schwartz2020green,xu2021survey}.

Due to considerations of fair comparisons and computational resources, we did not try super-large models which use at least 1.9B or more parameters like BEITv3~\citep{wang2022image}, CoCa~\citep{yu2022coca} and PaLI~\citep{chen2022pali}. We also did not pre-train large size model on large-scale datasets. However, scalability is also an important factor for foundation models. We leave the investigations to future work.

\paragraph{Potential Risks.}
The image-text pairs use for training our model are mostly derived from lexical databases and image queries in English, resulting in source material with a North
American or Western European bias.

\section{Conclusion}


In this work, we address the problem of how to build a general foundation model that can perform the best for all the understanding tasks of language, vision, and vision-language. We propose a new method for training general foundation model with two new and effective techniques, bringing in {\ModelName}, to learn rich language, vision, and vision-language representations at the same time. Experimental results demonstrate that {\ModelName} outperforms other general foundation models by a large margin. Moreover, {\ModelName} can even be better than or comparable to the SOTA foundation models specifically designed for language, vision, or vision-language understanding tasks. 



\vspace{-0.3em}

\bibliography{anthology,custom}
\bibliographystyle{acl_natbib}

\newpage
\appendix

\begin{table*}[ht]
\centering
\resizebox{\textwidth}{!}{%
\begin{tabular}{l | ccc | ccccc | ccc | cccc}
\toprule
\multirow{2}*{Methods}  & \multicolumn{3}{c}{Multimodal data} & \multicolumn{5}{c}{Pretraining Objectives} & \multicolumn{3}{c}{Fusion Arch.} & \multicolumn{4}{c}{Target Modalities} \\
  \cmidrule(lr){2-4}
  \cmidrule(lr){5-9}
  \cmidrule(lr){10-12}
  \cmidrule(lr){13-16}
 & public & dataset(s) & size & Contr. & ITM & BBP & 
 (M/P)LM & Unimodal  & ST & CT & MT & V & CV\&L & MV\&L & L \\ 

\midrule
RoBERTa~\citep{liu2019roberta} & -- & -- & -- & -- & -- & -- & -- & MLM & -- & -- & -- & -- & -- & -- & \Checkmark \\
BEiTv2~\citep{peng2022beit} & -- & -- & -- & -- & -- & -- & -- & MIM & -- & -- & -- & \Checkmark & -- & -- & -- \\
X-VLM~\citep{zeng2021multi,zeng2022x} & \Checkmark & Combination & 5M & \Checkmark & \Checkmark & \Checkmark & MLM & -- & -- & \Checkmark & -- & -- & \Checkmark & \Checkmark & -- \\
VLMo~\citep{wang2021vlmo} & \Checkmark & Combination & 5M & \Checkmark & \Checkmark & -- & MLM & MLM+MIM & -- & -- & \Checkmark & -- & \Checkmark & \Checkmark & -- \\
CLIP~\citep{DBLP:conf/icml/RadfordKHRGASAM21} & \XSolidBrush & WebImageText & 400M & \Checkmark & -- & -- & -- & -- & -- & -- & -- & \Checkmark & \Checkmark & -- & -- \\ 
ALIGN~\citep{jia2021scaling} & \XSolidBrush & JFT & 1.8B & \Checkmark & -- & -- & -- & -- & -- & -- & -- & \Checkmark & \Checkmark & -- & -- \\ 
SimVLM~\citep{wang2021simvlm} & \XSolidBrush & JFT & 1.8B & -- & -- & -- & PrefixLM & PrefixLM & \Checkmark & -- & -- & $\ast$ & -- & \Checkmark & \Checkmark \\ 
CoCa~\citep{yu2022coca} & \XSolidBrush & JFT & 4.8B & \Checkmark & -- & -- & LM & -- & \Checkmark & -- & -- & \Checkmark & \Checkmark & \Checkmark & -- \\
UNIMO-2~\citep{li2021unimo} & \Checkmark & Combination & 5M & -- & \Checkmark & -- & MLM & VCL & \Checkmark & -- & -- & \Checkmark & \Checkmark & \Checkmark & \Checkmark \\
OFA~\cite{wang2022ofa} & \Checkmark & Combination & 15M & -- & -- & -- & LM & LM & \Checkmark & -- & -- & $\ast$ & -- & \Checkmark & \Checkmark \\
DaVinci~\citep{diao2022prefix} & \Checkmark & Combination & 46M & -- & -- & -- & PrefixLM + PrefixIM & PrefixLM & \Checkmark & -- & -- & \Checkmark & -- & \Checkmark & \Checkmark \\
FLAVA~\citep{singh2021flava} & \Checkmark & Combination & 70M & \Checkmark & \Checkmark & -- & MLM & MLM+MIM & \Checkmark & -- & -- & \Checkmark & \Checkmark & \Checkmark & \Checkmark \\
Uni-Perceiver-MoE~\citep{zhu2022uni} & \Checkmark & Combination & 116M & -- & \Checkmark & -- & LM+MLM & LM+MLM+Classify. & \Checkmark & -- & -- & \Checkmark & \Checkmark & \Checkmark & \Checkmark \\
{\bf \ModelName} & \Checkmark & Combination & 5M & \Checkmark & \Checkmark & \Checkmark & MLM+MIM & MLM+MIM & -- & \Checkmark & -- & \Checkmark & \Checkmark & \Checkmark & \Checkmark \\
\midrule
\multicolumn{16}{l}{\textit{Super-Large Models}} \\
\textcolor{gray}{Flamingo~\citep{alayrac2022flamingo}} & \textcolor{gray}{\XSolidBrush} & \textcolor{gray}{Combination} & \textcolor{gray}{2.2B} & \textcolor{gray}{--} & \textcolor{gray}{--} & \textcolor{gray}{--} & \textcolor{gray}{LM} & \textcolor{gray}{--} & \textcolor{gray}{\Checkmark} & \textcolor{gray}{--} & \textcolor{gray}{--} & \textcolor{gray}{--} & \textcolor{gray}{\Checkmark} & \textcolor{gray}{\Checkmark} & \textcolor{gray}{--} \\
\textcolor{gray}{BEiT-v3~\citep{wang2022image}} & \textcolor{gray}{\Checkmark} & \textcolor{gray}{Combination} & \textcolor{gray}{21M} & \textcolor{gray}{--} & \textcolor{gray}{--} & \textcolor{gray}{--} & \textcolor{gray}{MLM} & \textcolor{gray}{MLM+MIM} & \textcolor{gray}{--} & \textcolor{gray}{--} & \textcolor{gray}{\Checkmark} & \textcolor{gray}{$\ast$} & \textcolor{gray}{\Checkmark} & \textcolor{gray}{\Checkmark} & \textcolor{gray}{--} \\
\textcolor{gray}{PaLI~\citep{chen2022pali}} & \textcolor{gray}{\XSolidBrush} & \textcolor{gray}{WebImageText} & \textcolor{gray}{41B} & \textcolor{gray}{--} & \textcolor{gray}{--} & \textcolor{gray}{--} & \textcolor{gray}{LM} & \textcolor{gray}{--} & \textcolor{gray}{\Checkmark} & \textcolor{gray}{--} & \textcolor{gray}{--} & \textcolor{gray}{\Checkmark} & \textcolor{gray}{\Checkmark} & \textcolor{gray}{\Checkmark} & \textcolor{gray}{\Checkmark} \\
\bottomrule
\end{tabular}
}
\caption{\textbf{Comparison of recent foundation models in different modalities.} Contr. indicates contrastive learning. ITM is short for image-text matching. BBP represents boundary box prediction.
(M/P)LM means image-conditioned (masked/prefix) language modeling. 
V, CV\&L, MV\&L and L stand for vision tasks, cross-modal retrieval tasks, multi-modal fusion tasks and language tasks respectively. 
ST, CT and MT are abbreviations for single Transformer, cross-attention Transformer and multiway Transformer.
VCL stands for visual contrastive learning.
$\ast$ means the modality is partially targeted (SimVLM and OFA include ImageNet.).
Giant models with over 1B parameters (e.g. BEiT-3) are in grey since they are not directly comparable with other models.
}
\label{tab:model_comparisions}
\end{table*}

\section{Comparison of Foundation Models}
\label{appendix:model_comparisions}
Table~\ref{tab:model_comparisions} shows an extensive comparison of
recent foundation models and {\ModelName} on multiple axes. Previous work either (i) perform best on uni-modal tasks~\citep{liu2019roberta,peng2022beit} or vision-language tasks~\citep{zeng2021multi,zeng2022x}; (2) target a specific uni-modal domain along with part of vision-and-language tasks~\citep{wang2021vlmo,DBLP:conf/icml/RadfordKHRGASAM21,jia2021scaling,wang2021simvlm,yu2022coca,wang2022ofa,diao2022prefix}; or (3) target all domains but cannot perform best on all the tasks~\citep{li2021unimo,singh2021flava,zhu2022uni}. Our model, {\ModelName}, is a general foundation model that can perform the best for all the understanding tasks of language, vision, and vision language. 



\section{Details of Pre-training Datasets}
\label{appendix:details_of_datasets}

We conduct our experiments on several widely used public datasets, consisting of two in-domain datasets, COCO~\cite{lin2014microsoft} and Visual Genome (VG)~\cite{krishna2016visual}, and two out-of-domain datasets, SBU Captions~\cite{ordonez2011im2text} and Conceptual Captions (CC)~\cite{sharma2018conceptual}. Following X-VLM~\citep{zeng2021multi,zeng2022x}, we use annotations of objects and regions from RefCOCO~\cite{yu2016modeling}, Objects365~\cite{shao2019objects365} and OpenImages~\cite{kuznetsova2018open}. We also include uni-modal data, RoBERTa corpus~\citep{liu2019roberta}, $C4$ datasets~\citep{raffel2020exploring} and Imagenet21K~\citep{ridnik2021imagenet21k}.

For our ``more data" setting, we scale up the pre-training dataset by including image-text pairs from Conceptual 12M dataset (CC-12M)~\cite{changpinyo2021conceptual} and LAION~\cite{schuhmann2022laion}. Thanks to LAION, we can use a large-scale public corpus of image-text pairs. However, we note that there are amounts of ``low-quality" image text pairs, as it is only filtered by the CLIP score. The clip score is deceptive when an image contains word tokens in its caption. Therefore, we apply three filters, OCR filter, text filter, and image filter, to capture ``high-quality" image-text pairs from LAION. Note that we only use English data in LAION.
The OCR filter will remove an image (image-text pair) when its OCR text contains more than four words or any token in the caption.
The text filter will remove a text image (image-text pair) if it is an address or contains only digits or symbols. The image filter will remove an image (image-text pair) if the shorter edge is smaller than 224 pixels, and also remove an image (image-text pair) if the height/width or width/height ratio is greater than 3. Finally, we have 1.3B paired data after all three filters. Statistics of the pre-training datasets are shown in Table~\ref{tab:data}.

\begin{table}[ht]
\small
\centering	
\resizebox{0.98\columnwidth}{!}{
\begin{tabular}	{ l | l |  l | l | l}
\toprule
Dataset & \# Images & \# Texts & \# Objects & \# Regions \\
\midrule
COCO & 0.11M & 0.55M & 0.45M & -\\
VG & 0.10M & - & 2.0M & 3.7M \\
SBU & 0.86M & 0.86M & - & -\\
CC-3M & 2.9M & 2.9M & - & - \\
Objects365 & 0.58M  & - & 2.0M & - \\ 
OpenImages & 1.7M  & - & 4.2M & - \\
C4 & - & 800GB & - & - \\ 
RoBERTa Corpus & - & 160GB & - & - \\ 
ImageNet-21k & 14M  & - & - & - \\
\midrule
\multicolumn{5}{l}{\textit{More Data}} \\
CC-12M & 11.1M & 11.1M & - & - \\
LAION & 1.3B & 1.3B & - & - \\
 \bottomrule
\end{tabular}}
\vspace{0.2cm}
\caption
{
Statistics of the pre-training datasets. 
}
\label{tab:data}
\end{table}

\section{Details of Downstream Tasks}
\label{appendix:details_of_downstream_tasks}

We report overall performance on eight language tasks from GLUE~\cite{wang2018glue}, eight vision tasks following OmniVL~\cite{wang2022omnivl} (More image classification tasks can be found in Table~\ref{tab:image_classifications}.), four multi-modal tasks, which are text-image retrieval on MSCOCO and Flickr, visual question answering (VQA~\cite{goyal2017making}), visual reasoning (NLVR2~\cite{suhr2019corpus}) and visual grounding (RefCOCO+~\cite{yu2016modeling}). For image-text retrieval task, we report both zero-shot results and fine-tuned results. For the ImageNet classification task, we report both linear evaluation results and fine-tuning results. The other vision tasks are evaluated in the linear evaluation setting. All the other tasks are evaluated in the fine-tuning setting.
Because the image resolution differs between pre-training and fine-tuning, the position parameters are adapted using linear interpolation.
For all downstream tasks, we apply random resize crops and horizontal flips augmentation for the images during training. More details of network architectures and hyper-parameters setups are given in Appendix \ref{appendix:details_of_hyper_params}.

\paragraph{Language Understanding. \newline} 
We conduct experiments on GLUE benchmark including MNLI~\citep{williams2018broad}, CoLA~\citep{warstadt2019neural}, MRPC~\citep{dolan2005automatically}, QQP~\citep{iyer2017first}, SST-2~\citep{socher2013recursive}, QNLI~\citep{rajpurkar2016squad}, RTE~\citep{dagan2005pascal,haim2006second,giampiccolo2007third,bentivogli2009fifth}, and STS-B~\citep{agirre2007proceedings}. 
We follow the practice of BERT~\citep{devlin2018bert, liu2019roberta} and feed the input into the language encoder, and the hidden state of the \texttt{[CLS]} is fed into a new multi-class linear classifier or regression head.

\paragraph{Vision Understanding. \newline}
We conduct vision experiments on both fine-tuning and linear evaluation (linear eval).
The linear evaluation follows a common practice~\citep{caron2021emerging, he2020momentum, singh2021flava} in self-supervised learning to evaluate the representation quality, where the pre-trained backbone model is frozen, and an MLP head is appended on top of it.
We choose 7 popular datasets following OmnVL~\citep{wang2022omnivl}:
ImageNet~\citep{russakovsky2015imagenet}, Food101~\citep{bossard2014food}, CIFAR10~\citep{krizhevsky2009learning}, CIFAR100~\citep{krizhevsky2009learning}, DTD~\citep{cimpoi2014describing}, Pets~\citep{parkhi2012cats} and  Flowers102~\citep{nilsback2008automated}.

\paragraph{Vision-Language Understanding. \newline} 
\noindent \textbf{Image-Text Retrieval} We evaluate {\ModelName} on both MSCOCO and Flickr30K datasets. 
We adopt the widely used Karpathy split~\citep{karpathy2015deep} for both datasets.
Following the previous work~\citep{li2021align,zeng2021multi,zeng2022x}, we first encode images and texts separately and calculate $s(I,T)$ to obtain the top-$k$ candidates, and then use the fusion encoder to re-rank the candidates. 

\noindent \textbf{Visual Question Answering} The task requires the model to predict an answer given an image and a question. We evaluate {\ModelName} on the VQA v2.0 dataset~\cite{goyal2017making}. Following the previous work~\citep{zeng2021multi}, we use a Transformer decoder to generate answers based on the outputs of the fusion module. The decoder network shares the same network architecture with the fusion encoder. Note that we use an image resolution of 768*768 for the final result of {\ModelNameB}, and use an image resolution of 480*480 for {\ModelNameB} in ablation studies for efficient fine-tuning.

\noindent \textbf{Visual Reasoning} We evaluate {\ModelName} on a widely used benchmark NLVR2~\cite{suhr2018corpus}. The task allows the model to determine whether a text describes the relations between two images. Following previous work~\cite{wang2021vlmo, bao2022vl}, we formulate the triplet input into two image-text pairs, each containing the text description and an image. We then concatenate the final output [CLS] features of the fusion module of the two pairs to predict the label.

\noindent \textbf{Visual Grounding} We evaluate {\ModelName} on RefCOCO+~\cite{yu2016modeling}. Given an image and a text description as input, the final output $[CLS]$ features of the fusion module is utilized to predict the bounding
box $(cx, cy, w, h)$, i.e. the normalized center coordinates, width, and height.

\begin{table}[t]
\centering
\small
\resizebox{0.98\columnwidth}{!}{
\begin{tabular}{lccccccccc}
\toprule
\multirow{2}{*}{Model}& \multicolumn{2}{c}{Param} & \multirow{2}{*}{Hidden} & \multicolumn{3}{c}{Layers} \\
& Total & Trans. & & Vision & Text & Fusion \\
\midrule
{\ModelNameB} & 327M & 284M & 768 & 12 & 12 & 12 \\
\bottomrule
\end{tabular}}
\caption{Size variants of {\ModelName}. All modules consist of transformer layers. Param indicates the parameter. Total means the total parameter number, and Trans. indicates the parameter number for Transformer layers.}
\label{tab:modelsize}
\end{table}

\section{Details of hyper parameters}
\label{appendix:details_of_hyper_params}

\paragraph{Pre-training} 
{\ModelNameB} is implemented with a 12-layer language encoder, a 12-layer vision encoder, and a 12-layer fusion encoder, 768 dimensions for hidden states, 3072 for intermediate size, and 128 for maximum input length.
We initialize the language encoder with RoBERTa and the vision encoder with BEiTv2.
The weight decay is set to 0.01 with $\beta_1 = 0.9, \beta_2 = 0.98$.
The learning rate is 1e-4 with a warm-up period for the first 2500 steps and then linearly decayed to 0.
In each batch, there are 3072 image-text pairs, 3072 images, and 8192 text-only sentences.
We use center-crop to resize each image to the size of 224×224. The model sizes and default hyper-parameter settings are shown in Table~\ref{tab:modelsize} and Table~\ref{tab:impl_pretrain}, respectively.

\begin{table}[ht]
\centering
\resizebox{0.9\linewidth}{!}{
\begin{tabular}{c|c}
\toprule
config & value \\
\hline
optimizer & AdamW \\
learning rate & 1e-4 \\
weight decay & 0.01 \\
optimizer momentum & $\beta_1, \beta_2{=}0.9, 0.999$ \\
language batch size & 8192 \\
vision batch size & 3072 \\
vision-language batch size & 3072 \\
learning rate schedule & linear decay \\
warmup steps & 2500  \\
training steps & 200k  \\
augmentation & RandomResizedCrop \\
image res & 224*224 \\
patch size & 16 \\
text length for MLM & 128 \\
text length for IMLM & 30 \\
\bottomrule
\end{tabular}}
\caption{Pre-training setting.}
\label{tab:impl_pretrain} 
\end{table}

\paragraph{Fine-tuning}
The learning rate is $\in$ \{1e-5, 2e-5, 5e-5\} and our model is optimized by AdamW.
Because the image resolution differs between pre-training and fine-tuning, the position parameters are adapted using linear interpolation.
For all downstream tasks, we apply random resize crops and horizontal flips augmentation during training.
The default settings for text classification, image classification and vision-language understanding are shown in Tables~\ref{tab:impl_glue}, \ref{tab:impl_linear}, \ref{tab:impl_imagenet} and \ref{tab:impl_finetune}, respectively. Note that the resolution for VQA is different as described in Section~\ref{appendix:details_of_downstream_tasks}.

\begin{table}[ht]
\resizebox{0.9\linewidth}{!}{
\begin{tabular}{c|c}
\toprule
config & value \\
\hline
optimizer & AdamW \\
learning rate & \{1e-5, 2e-5, 5e-5\} \\
weight decay & 0.0 \\
optimizer momentum & $\beta_1, \beta_2{=}0.9, 0.999$ \\
batch size & \{16, 32, 64\} \\
learning rate schedule & linear decay \\
warmup ratio & 0.0 \\
training epochs & \{5, 10, 20\} \\
\bottomrule
\end{tabular}}
\caption{Text classification: GLUE setting. 
\label{tab:impl_glue}}
\end{table}

\begin{table}[ht]
\resizebox{0.9\linewidth}{!}{
\begin{tabular}{c|c}
\toprule
config & value \\
\hline
optimizer & AdamW \\
learning rate & [2e-5, 4e-5] \\
weight decay & 0.01 \\
optimizer momentum & $\beta_1, \beta_2{=}0.9, 0.999$ \\
batch size & [256, 2048] \\
learning rate schedule & linear decay \\
warmup rate & 0.1 \\
training epochs & 100 \\
augmentation & RandomResizedCrop \\
image res & 224*224 \\
patch size & 16 \\
\bottomrule
\end{tabular}}
\caption{Image classification: Linear probing setting. 
\label{tab:impl_linear}}
\end{table}

\begin{table}[ht]
\resizebox{0.9\linewidth}{!}{
\begin{tabular}{c|c}
\toprule
config & value \\
\hline
optimizer & AdamW \\
learning rate & 4e-5 \\
minimal learning rate & 1e-7 \\
weight decay & 0.01 \\
optimizer momentum & $\beta_1, \beta_2{=}0.9, 0.999$ \\
batch size & 1024 \\
learning rate schedule & linear decay \\
warmup rate & 0.1 \\
training epochs & 100 \\
augmentation & RandomResizedCrop \\
image res & 224*224 \\
patch size & 16 \\
label smoothing  & 0.1 \\
mixup prob. & 1.0 \\
cutmix prob. & 1.0 \\
\bottomrule
\end{tabular}}
\caption{ImageNet classification: Fine-tuning setting. 
\label{tab:impl_imagenet}}
\end{table}

\begin{table}[ht]
\resizebox{0.9\linewidth}{!}{
\begin{tabular}{c|c}
\toprule
config & value \\
\hline
optimizer & AdamW \\
learning rate & \{1e-5, 2e-5, 5e-5\} \\
weight decay & 0.01 \\
optimizer momentum & $\beta_1, \beta_2{=}0.9, 0.999$ \\
batch size & \{64, 192, 512\} \\
learning rate schedule & linear decay \\
warmup rate & 0.1 \\
training epochs & \{10, 15, 20\} \\
augmentation & RandomResizedCrop \\
image res & 384*384 \\
patch size & 16 \\
\bottomrule
\end{tabular}}
\caption{Vision-Language understanding: fine-tuning setting.}
\label{tab:impl_finetune} 
\end{table}


\end{document}